\documentclass[runningheads]{llncs}

\usepackage{eccv}

\usepackage{eccvabbrv}

\usepackage{graphicx}
\usepackage{booktabs}
\usepackage{adjustbox}
\usepackage{makecell}
\usepackage{multirow}
\usepackage{pifont}
\usepackage{subcaption}
\usepackage{blindtext}
\usepackage{enumitem}
\usepackage[accsupp]{axessibility}  

\usepackage{hyperref}

\usepackage{orcidlink}
\newcommand{\backronym}{LEIA\xspace}
\newcommand{\xmark}{\ding{55}}%
\newcommand{\x}{\mathbf{x}}
\newcommand{\col}{\mathbf{c}}
\newcommand{\dir}{\mathbf{d}}

\begin{document}

\title{LEIA: Latent View-invariant Embeddings for Implicit 3D Articulation} 


\author{Archana Swaminathan\inst{1}\orcidlink{0009-0005-1412-9790} \and
Anubhav Gupta\inst{1}\orcidlink{0000-0003-4017-2862} \and
Kamal Gupta\inst{1}\orcidlink{0000-0003-4058-9991} \and 
Shishira R Maiya \inst{1}\orcidlink{0000-0002-5346-9510}
\and
Vatsal Agarwal\inst{1}\orcidlink{0009-0004-9470-198X}
\and 
Abhinav Shrivastava\inst{1}\orcidlink{0000-0001-8928-8554}}

\authorrunning{A. Swaminathan et al.}

\institute{University of Maryland, College Park MD 20742, USA \\
\url{https://archana1998.github.io/leia/}}
\maketitle
\begin{abstract}
Neural Radiance Fields (NeRFs) have revolutionized the reconstruction of static scenes and objects in 3D, offering unprecedented quality. However, extending NeRFs to model dynamic objects or object articulations remains a challenging problem. Previous works have tackled this issue by focusing on part-level reconstruction and motion estimation for objects, but they often rely on heuristics regarding the number of moving parts or object categories, which can limit their practical use. In this work, we introduce \backronym, a novel approach for representing dynamic 3D objects. Our method involves observing the object at distinct time steps or ``states'' and conditioning a hypernetwork on the current state, using this to parameterize our NeRF. This approach allows us to learn a view-invariant latent representation for each state. We further demonstrate that by interpolating between these states, we can generate novel articulation configurations in 3D space that were previously unseen. Our experimental results highlight the effectiveness of our method in articulating objects in a manner that is independent of the viewing angle and joint configuration. Notably, our approach outperforms previous methods that rely on motion information for articulation registration. 
\end{abstract}

\section{Introduction}
\label{sec:intro}

Our world is full of dynamic objects moving and interacting in space and time. Humans develop this (rather impressive) understanding of how these everyday objects move and interact in three-dimensional space at a very early stage of the brain development~\cite{saffran1996statistical}. The task involves understanding not only the static geometry but also the dynamic movements and spatial relationships between parts of an object, often referred to as articulations of the object. Understanding and representing object articulations from images and/or videos is also pivotal in enabling machines to perceive and navigate the physical world with finesse. In this work, we propose a novel method to model the object articulations by learning view-invariant latent embeddings of the 3D object from multiview images.
\begin{figure}[t!]
    \centering
    \includegraphics[trim=1cm 5.5cm 1cm 2cm, clip, width=\linewidth]{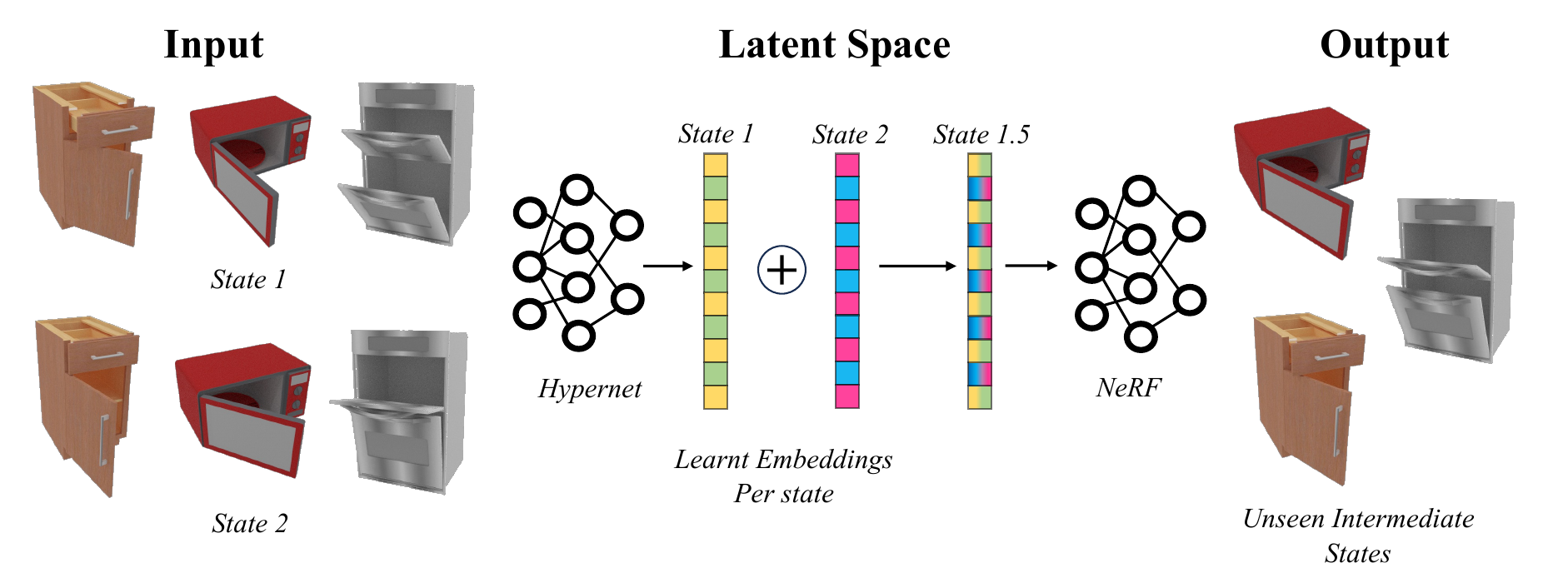}
    \caption{Our method \textbf{\backronym}, takes in multi-view images of an object in four articulation states and is able to learn a view-invariant latent embedding for the state. We show that we can interpolate between the latents to generate any number of intermediate unseen states for the object using \backronym, given the camera position.}
    \label{fig:teaser}
\end{figure}
Existing works in modeling object articulation in the three dimensional space typically use priors in the form of pre-trained large scale models~\cite{yao2022lassie}, videos~\cite{park2021hypernerf}, or assumptions regarding rigidity/shape of the object~\cite{jiayi2023paris}. These approaches often fail to generalize for the long-tail of objects, especially in the case when a video of the object with articulation is unavailable, or in presence of large and multiple articulations in the object.
In our work, we postulate that multiview images of an object in different states can provide enough signal to model its 3D shape and articulation, even without priors from large-scale foundation models or information from videos. We achieve this by learning a generalizable, view-invariant latent embedding of the object for different states. We train these embeddings jointly with a hypernetwork that predicts weights of a NeRF \cite{mildenhall2020nerf}  parameterizing the object state. The hypernetwork can be trained using multi-view images of discretized states of the object, each state representing a different object articulation. In \Cref{fig:teaser}, we show three objects from our dataset in two different articulation states. The latent embeddings representing each state or articulation can be interpolated during test time to generate the weights of a NeRF that is able to reconstruct a never seen before state of the object.

Our key contributions can be summarized as following:
\begin{itemize}[topsep=0em]
\item We introduce an end-to-end method \backronym for generating novel states for articulated objects solely with multiview images captured at multiple states.
\item We demonstrate that interpolating between embeddings can generate states of articulations of object not seen during training. The embedding space becomes interpolable with a manifold loss that encourages the latents to follow a structure that establishes a linear relationship between them, by minimizing the distance between the nearest neighbours in the latent space.  
\item Remarkably, \backronym achieves this without the need for any ground-truth 3D supervision, motion information, or articulation codes, establishing its versatility and effectiveness in capturing complex articulations.
\item Our analyses demonstrate \backronym's robustness to single and multiple articulations, as well as combinations of motions. We can disentangle articulations in different object parts if multiple are movable, making it scalable without constraints on the number of parts or motion types, unlike prior work.
\end{itemize}

\section{Related Work}

\noindent\textbf{Neural Radiance Fields.}
Neural Radiance Fields \cite{mildenhall2020nerf, mueller2022instant,wang2021neus, Barron_2022_CVPR} have proven to be a massive success in modeling 3D scenes, due to the high fidelity of 3D reconstruction and novel view synthesis for static scenes.
For dynamic scenes, Neural Volumes \cite{Lombardi:2019} utilizes an encoder-decoder voxel-based representation, complemented by an implicit voxel warp field. Occupancy Flow \cite{Niemeyer_2019_ICCV} tackled non-rigid geometry by assigning a motion vector to each point in both space and time, but requires full 3D ground truth supervision. Some of the first dynamic NeRF approaches \cite{pumarola2020d, tretschk2021nonrigid, park2021nerfies} optimize an underlying volumetric deformable function and \cite{li2022neural} conditions the NeRF on time. \cite{xian2021space, park2021nerfies,lin2022efficient} followed this work, by learning a 5D spatiotemporal neural field. However, the approaches above usually require a video as input and do not handle well the case of large articulations in everyday objects given multiple states of objects as input.

\noindent\textbf{3D Representations for Articulation.}
Due to the fine-grained nature of the task, deep neural network models for representing articulation require conditioning or refining parts of the architecture to suit the task. Category-specific reconstruction of deformable objects from images \cite{cmrKanazawa18, ucmrGoel20, kulkarni2020articulation, kokkinos2021to}, sparked interest in identifying and recovering the deformation in the 3D space \cite{kulkarni2020articulation, Neverova2020ContinuousSurfaceEmbeddings, yao2022-lassie}.  
Several works focus on shape reconstruction from videos \cite{yang2021lasr, yang2021viser, yang2022banmo, tan2023distilling} that estimate a shape template for humans and animals. Other works that learn articulation from videos include Qian et al. \cite{Qian22} that detects and segments articulation planes from in-the-wild videos and \cite{NEURIPS2022_d2cc447d, yuan2021star} which decouple the static and dynamic parts of videos.
These works focus on modeling humans and animals where a large amount of data is available. 
Preliminary work in using images and shape started with A-SDF \cite{pumarola2020d}, which generates unseen articulations using Signed Distance Function \cite{Park_2019_CVPR}, by providing an input shape and articulation code. CaDeX \cite{Lei_2022_CVPR} is a unified representation for shape and motion, obtained from a point cloud input. Ditto \cite{jiang2022ditto} is a similar work that uses implicit representations for joint geometry and articulation modeling, with ground truth point clouds. CLA-NeRF \cite{tseng2022clanerf} learns unseen articulated states from observing multi-view image input along with articulation information. Wei et. al \cite{wei2022nasam} obtains an SDF-based articulable representation of common objects by feeding in images of articulated states across multiple categories, and CARTO \cite{heppert2023carto} uses stereo images as input and uses a stereo encoder to infer the 3D shape, 6D pose etc. of multiple unknown objects. \cite{Chu_2023_CVPR} focuses on manipulating object shapes to deform according to specified articulation commands. Moving away from 3D supervision and articulation annotations, Jiayi et al. \cite{jiayi2023paris} proposed PARIS, a method that is able to obtain unseen articulated states of an object, given multi-view images of just the start and end state of articulation. PARIS employs a composite rendering based approach, by decoupling the object into a static and moving part and then separately estimating and compositing the static and mobile neural radiance fields. In our work, we start \backronym with a similar input setting, by preparing multi-view renderings of objects in different articulation states. Unlike PARIS, we do not decouple the object into static and moveable parts, we rather learn to predict the weights of a single NeRF for any unseen articulated state using a state-modulated HyperNetwork. We aren't limited by the decoupling and the learned motion prior in the case of PARIS, which enables us to scale and learn any amount and kind of motion an object can possibly have, including combinations. We show a summed comparison of \backronym with prior work in \Cref{tab:relatedwork}.
\begin{table}[t!]
    \centering
    \footnotesize
    \caption{\textbf{Comparison of \backronym with existing methods.} We show that \backronym is the first approach that does not use or learn any explicit prior along with not having any articulation input. This gives us flexibility in scaling to modeling articulations of objects with more than one part, and can thus handle a wide range of motion. We also have one universal model that can learn to represent both prismatic and revolute motion, unlike PARIS that has two separate models. 3D Sup. refers to ``3D Supervision''.}
    \begin{adjustbox}{width=\columnwidth,center}
    \renewcommand{\tabcolsep}{2pt}
    \begin{tabular}{@{}llllllrr@{}}
    \toprule
    Method & Image/Shape Input & Dataset & Articulation Input &  3D Sup.? & Prior & \# States & \# Parts \\
    \midrule
       A-SDF~\cite{pumarola2020d}  & Shape Code & Shape2Motion & Articulation Code & \checkmark & Articulation & $\geq4$  & $\geq1$\\
       CLA-NeRF\cite{tseng2022clanerf} & Multiview Images & PartNet-Mobility & Articulated Pose & \xmark & Articulation & $\geq4$ & 1 \\
       NASAM\cite{wei2022nasam} & Multiview Images++ & PartNet-Mobility & N/A & \xmark & Category & $\geq4$ & $\geq1$ \\
       CARTO\cite{heppert2023carto} & Stereo Images & \emph{Custom} & Joint Code & \checkmark & Articulation & $\geq4$ & 1 \\
       Ditto\cite{jiang2022ditto} & Point Clouds & Shape2Motion & Annotations & \checkmark & Articulation & 2  & 1 \\
       PARIS\cite{jiayi2023paris} & Multiview Images & PartNet-Mobility & N/A & \xmark & Motion (learned) & 2 & 1 \\
       \midrule
       \textbf{\backronym} & \textbf{Multiview Images} & \textbf{PartNet-Mobility} & \textbf{N/A} & \textbf{\checkmark} & \textbf{No} & \textbf{4} & $\mathbf{\geq1}$ \\
       \bottomrule
    \end{tabular}
    \end{adjustbox}
    \label{tab:relatedwork}
\end{table}

\noindent\textbf{Hypernetworks with Implicit Neural Representations.}
Hypernetworks, a specialized class of networks designed to predict parameters for another network, aim to achieve generalization across novel tasks \cite{ha2017hypernetworks}. In the realm of stylizing 3D scenes, \cite{chiang2022stylizing} employed a hypernetwork for applying diverse styles, while scene reconstructions from limited data points were achieved by \cite{sitzmann2019scene,sitzmann2021light}. Despite the promise shown in representation, these hypernetworks operate on input data points, necessitating test-time optimizations and rendering them unsuitable for compression tasks. Rather than using the provided data (image/video) as input for the hypernetwork, an alternative approach involves employing an auto-decoder framework. In this framework, a learnable latent, without the need for an encoder, represents a data point. This technique, applied by \cite{sen2022inr} to represent a dataset of videos, assigns each latent to a distinct video. While this method yields a representation for each set of frames, the lack of decoupling in spatial-temporal coordinates limits its scalability to real-world frames. On a related note, \cite{sen2023hyp} extended a similar approach to 3D shapes and scenes, effectively acquiring a latent representation suitable for various downstream tasks.
We use a similar framework for learning the latents corresponding to each articulated state in our method.

\section{Method}
\noindent\textbf{Background.}
Our architecture is based on neural radiance fields, or NeRF~\cite{mildenhall2020nerf}, which parameterizes the radiance and volume density at a 3D location of a scene as observed from a camera placed in a particular position using a neural network. 
$F_\text{NeRF}: (\x, \dir) \xrightarrow \; (\col, \sigma)$, 
where $\col = (r,g,b)$, and $\sigma$ represent the radiance and volume density, and $\x$ and $\dir$ represent the 3D location and viewing direction respectively. To render a pixel, the radiance $C(\mathbf{r})$ of a camera ray $\mathbf{r}(s) = \mathbf{o} + s\cdot\mathbf{d}$ is integrated from near to far bounds $s_n$ and $s_f$ such that,    $C(\mathbf{r}) = \int_{s_n}^{s_f} T(s)\sigma(\mathbf{r}(s))\mathbf{c}(\mathbf{r}(s), \mathbf{d}) \, ds,$
where the function $T(s)$ denotes the accumulated transmittance from $s_n$ to $s$. To optimize the parameters of the MLP, a loss function is used that measures the discrepancy between the ground-truth and rendered images. Traditionally, the L2 loss is used for this purpose. However, to make the training more robust to outliers and to improve convergence, the Smooth L1 Loss can be employed as an alternative. The Smooth L1 Loss is a combination of L1 and L2 losses, behaving like L1 loss for large errors and like L2 loss for small errors. The loss function for the NeRF model using the Smooth L1 Loss is defined as:
\begin{equation}
    \text{SmoothL1Loss}(x) = \begin{cases} 
0.5 \cdot x^2 & \text{if } |x| < 1 \\
|x| - 0.5 & \text{otherwise}
\end{cases}
\end{equation}

\begin{equation}
\begin{aligned}  
    L_\text{SmoothL1-NeRF} = \sum_{r \in R} \text{SmoothL1Loss}\left(\hat{C}(\textbf{r}) - C(\textbf{r})\right)
\end{aligned}
\end{equation}

\noindent where $R$ is the set of rays used for sampling, and $\hat{C}(\textbf{r})$ and $C(\textbf{r})$ represent the ground-truth and rendered colors, respectively. By minimizing this loss function, the NeRF model learns to accurately reproduce the radiance of the scene, leading to high-quality image synthesis from novel viewpoints. While this formulation works well for generating novel scene viewpoints for a static scenes, it cannot handle dynamic scenes. \cite{xian2021space} tried to address this shortcoming by making the MLP learn a spatiotemporal radiance field and used time, $t$, as an additional input. This works in principle but it becomes expensive to scale to lengthier videos due to constantly increasing sampling space for NeRF. We use the $L_{\text{SmoothL1-NeRF}}$ loss along with AdamW optimization to train \backronym, along with a $L_{\text{mask}}$ that is the BCE loss between the predicted opacity and ground truth foreground loss.

\begin{figure} [t!]
    \centering
    \includegraphics[trim=0cm 0cm 0cm 0cm, clip, width=\linewidth]{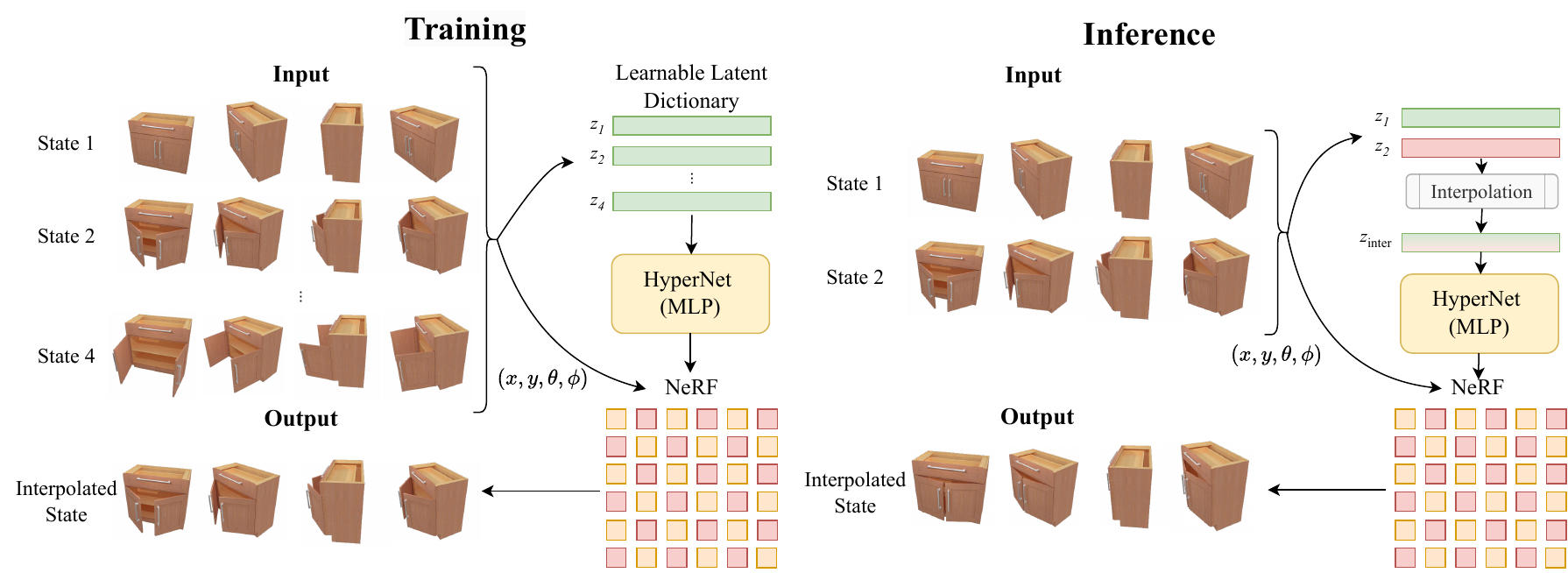}
    \caption{\textbf{Overview of our method.} We take multi-view images in different states as input. A learnable latent dictionary based off an autoencoder learns an embedding per state id. The latent embedding is used as an input to the hypernet, that modulates and generates weights of the NeRF to reconstruct the state that is fed in. At inference time, we do a weighed interpolation of the learnt latents to obtain a corresponding newly generated intermediate state.}
    \label{fig:method}
\end{figure}

\subsection{Approach}

Hypernetworks, or hypernets for short, are neural networks that generate weights for another neural network, known as the target network. Hypernets can be conditioned on various domains and can predict the weights for multiple neural networks simultaneously, if trained appropriately.

In this work, we use a hypernet to modulate a Neural Radiance Field (NeRF) based on the state of articulation, thereby learning a parametrization for each state. We employ a learnable latent embedding, $Z$, as input to this hypernet $h_{l}, l \in L$ where $L$ is the number of layers of the hypernet. This approach allows us to not only modulate the NeRF but also create useful representations for each state. Our system can thus be represented as follows:
\begin{equation}
\begin{aligned}
    F_{\theta_t}(\x,\dir) &=  (\col_t, \sigma_t) \\
    \theta_t &= h(z_t) ,   z_t \in Z \label{eq:hyper}
    \end{aligned}
\end{equation}

The latents $Z$, give us an additional ability to interpolate across seen states and generate novel states of the object, as we show in our work. The set of latents, $Z$, in \cref{eq:hyper} can be understood as a learnable dictionary where keys are the state ids. Formally, we represent this as 
\begin{equation}
    Z = \{\;t: \; z_t \; | \; t \in [0, 1, 2 ... T] \}
\end{equation}
where, $T$ is the total number of discretized states sampled for the object. Each of these states, $z_t$, is used as an input to the hypernet to get the parameterization for NeRF as shown in \Cref{fig:method}. Once parameterized, the NeRF is trained using multi-view images of the object taken from various camera angles, providing a comprehensive view of the object's articulation in the current state. We sample only one state per batch.

Directly predicting the weights $\theta$ of the base network $f_{\theta}$, using the hypernet $h_l$, is expensive, parameter-heavy, and unsuitable for compression. Hence, we follow \cite{skorokhodov2021adversarial,schwarz2023modality} and instead predict low-rank matrices, which are then applied to the base network weights. This type of modulation acts as a form of subnetwork selection, analogous to systems proposed in \cite{frankle2018lottery,ramanujan2020s}. For a base network $f_{\theta}$ with $L$ layers, our formulation now looks like 
\begin{equation}
\begin{aligned}
    f_{\theta}((\x,\dir)|\theta^{l_{1}}_{t},\theta^{l_{2}}_{t} ... \theta^{l_{L}}_{t}) &= \col_{t}, \sigma_t \\
    \theta^{l}_{t} &= \eta ( P^{l} \times Q^{l}) \circ \theta^{l}\\
    h_{l}(z_{t}) &= [P^{l},Q^{l}]
\end{aligned}
\end{equation}
where $\theta^{l}$ represents the weights of the $l$-th layer and $\theta^{l}_{t}$ denotes the modulated weights for frame $t$.  
Here, $\eta$ signifies an activation function on the matrix-product of low rank matrices $\mathbf{P}^{l} \in \mathbb{R}^{K \times r}$ , $\mathbf{Q}^{l} \in \mathbb{R}^{r \times K}$, where $K$ is the width of the base network $f_{\theta}$ and rank $r \ll K$. 
These matrices are responsible for adjusting the weights $\theta_{l}$ as dictated by the corresponding hypernetwork $h_{l}$. Note that all hypernetworks use the same latent $z_{t} \in \mathbb{R}^{D}$  as input. The rank $r$  acts as a hyperparameter that controls the compression-performance trade-off. We further elaborate on details about model architecture and design choices for the hypernet, NeRF and the learnable latent dictionary in the supplementary.

\subsection{Interpolation}
Given an object with states $t_1$ and $t_2$, and a scale $\alpha$, the task of state interpolation involves creating $\alpha - 1$ coherent states between the given states. In order to achieve this, we do a linear interpolation on the state latents $z_1$ and $z_2$ and pass the resulting latent through the hypernetwork. This gives us the weight modulation required in the NeRF, and the updated base network is used to obtain different viewpoints of the intermediate state. 
\begin{equation}
\begin{aligned}
    z_{\text{inter}} &= (1 - \beta_i) \cdot z_{t} + \beta_i \cdot z_{t-1} \\
    \col_{\text{inter}}, \sigma_{\text{inter}} &= f_{\theta_{\text{inter}}}(\x, \dir; h(z_{\text{inter}}))
\end{aligned}
\end{equation}

\begin{equation}
   \text{where,  } \beta_i \in \left[\frac{1}{\alpha}, \frac{2}{\alpha}, . . ., \frac{\alpha - 1}{\alpha}\right]
\end{equation}
essentially generating $\alpha - 1$ states between any two given states. In our experiments, we quantitatively evaluate the result of the interpolation with unseen ground truth obtained from our dataset of all states, by linearly averaging the two extreme states present in the training, to obtain a latent for a state id.

\subsection{Model Architecture} 

In our experiments, both the base network $f_{\theta}$ and hypernetworks $h_{l}$  are simple MLPs that take in a coordinate input, and an id depicting the state of articulation.  A network $l_n$ is used to learn the latent dictionary for the states, which is a linear layer learnt in conjuction with the hypernet, and uses the \texttt{torch.nn.Embedding} class as a lookup table for the learnable latent embeddings, depicted by $z_t$ for $t \in T$. This latent is fed to the hypernetwork $h_l$, that modulates the weights of the base network $f_{\theta}$ for reconstructing the output of the corresponding state id $t$. The base network $f_{\theta}$ is based off the NeRF architecture of Instant-NGP \cite{mueller2022instant}. This architecture has separate fully-connected blocks for the geometry and texture of the scene, and we use separate hypernets to modulate each layer of these two FC blocks.

\subsection{Optimization: Loss functions and regularizers}
\noindent \textbf{Latent Manifold Loss.}
We use the latent manifold loss function that enforces a structured latent space by encouraging local consistency among learnt latents. For each latent vector, the loss computes the average Euclidean distance between the vector and its $K$ nearest-neighbours on the manifold. This process enforces a smooth and continuous latent manifold, which is beneficial for models that rely on meaningful linear interpolations between the points on the manifold, and for tasks where the geometry of the latent space is crucial. Mathematically, we represent this loss for a particular state-id \textit{i} as:
\begin{equation}
\mathcal{L}_{\text{manifold}}(\mathbf{l}_i) = \frac{1}{K} \sum_{k=1}^{K} \left\| \mathbf{l}_i - \mathbf{n}_{k} \right\|_2^2
\end{equation}
where \( \mathbf{n}_{k} \) are the \( K \) nearest neighbors of \( \mathbf{l}_i \) in the latent space, and \( \left\| \cdot \right\|_2 \) denotes the Euclidean (L2) norm. The loss is averaged over the selected latents and their nearest neighbors to ensure local uniformity in the manifold's geometry, where \( K \) is a hyperparameter that determines the number of neighbors considered.

\noindent\textbf{Depth and Occlusion Regularization.} Our depth and occlusion regularizations are designed to refine the clarity of rendered images by addressing occlusion and depth smoothness~\cite{Niemeyer2021Regnerf,yang2022freenerf}. The occlusion regularization loss, $L_{\text{occ}}$, aims to mitigate the obscuring of objects located beyond a specified depth threshold during rendering. This is accomplished by generating a binary mask, $m_k$, where the mask elements are set to 1 up to a certain index $M$ reflecting the regularization range, and 0 thereafter. The occlusion loss is then articulated as the normalized sum of the product of the mask and the sampled density values $\sigma_k$ along a ray, as given by 
\begin{equation}
    L_{\text{occ}} = \frac{1}{K} \sum_{k=1}^{K} \sigma_{k} \cdot m_{k},
\end{equation}
where $K$ is the total number of sampled points on the ray. This formulation drives the model to prefer representations that reduce occlusions close to the camera, ensuring objects further away are not improperly concealed.

For depth continuity, the depth smoothness regularization loss, \( L_{\text{DS}}(\theta, \mathbf{R}) \), enforces the gradual transition of depth values among neighboring pixels, reducing sharp depth disparities that cause visual inconsistencies. If a ray \( r \) intersects with a single depth value, this value is evaluated directly. In contrast, for multiple depth values, the loss is the aggregate of squared differences between adjacent depth estimates. Formally, for depth predictions \( \hat{d}_0(r) \), the loss is quantified as
\begin{equation}
L_{\text{DS}}(\theta, \mathbf{R}) = \sum_{r \in \mathbf{R}} \sum_{i,j=1}^{\text{Patch}-1} \left( \hat{d}_0(r_{ij}) - \hat{d}_0(r_{(i+1)j}) \right)^2 + \left( \hat{d}_0(r_{ij}) - \hat{d}_0(r_{i(j+1)}) \right)^2.
\end{equation}
This measure is averaged across the rays to yield a measure of the depth map's smoothness. By integrating these regularizers into our training regimen, we significantly dampen the disturbances introduced by overlapping and obscured pixels, resulting in a more consistent interpolation between states. These benefits are substantiated through extensive ablation studies and quantitative analyses presented in the experiments.

\noindent\textbf{Positional Encoding of the Latent.}
We incorporate positional encoding to capture the order of input elements, crucial for understanding articulation. Positional encoding injects the sequence with its inherent order, a key factor in articulation semantics. We employ the scheme from \cite{NIPS2017_3f5ee243}, which uses sine and cosine functions parameterized to encode varying frequencies. The computed positional encodings are added to the latent vector $z_t$, enriching it with semantic information that more accurately encodes the state of articulation. We analyze the effects of positional encoding on our model in the experimental section.
\section{Experiments}

\subsection{Setup}

\noindent\textbf{Datasets.}
In this work, we use the PartNet-Mobility dataset \cite{Xiang_2020_SAPIEN, Mo_2019_CVPR,chang2015shapenet}, a large-scale synthetic collection of articulated objects in over 40 categories.  The dataset has a variety of articulations defined, with objects comprising of single and multi-joint parts. Two types of motion, revolute (rotational) joint and prismatic (translation) articulation are represented in the dataset. To make up our training dataset, we choose 100 camera views that are arranged in a dome-like setup, capturing the upper hemisphere of the object. This is similar to the setup used by PARIS. We use the SAPIEN \cite{Xiang_2020_SAPIEN} library that the PartNet-Mobility dataset was released with, to render our RGB images at linearly spaced intervals of articulation to make up frames of a video that depicts the range of motion and obtain the camera parameters accordingly, converting it into the Blender~\cite{blender} coordinate system to fit in our codebase. We use instances from 8 different common household item categories \texttt{storage, microwave, laptop, oven, washer, dishwasher, sunglasses, box} and choose 1-4 objects per category, bringing our total number of objects to 12. We also show the efficacy of our method on 68 images from a real-world scene of a chest of drawers, which are captured with a mobile phone and post-processed to remove background. Throughout \backronym, we train with a total of four states and interpolate between the two extreme states.  
\begin{table}[!t]
  \centering
    \caption{\textbf{Quantitative Results for Interpolated State Reconstruction.} We compared our method with the PARIS baseline, trained on selected objects from the SAPIEN dataset. The results from three experiments are summarized below. The VanillaInt experiment involves simple interpolation of the latents. Our best-performing method, LEIA, introduces structure to the latents with a manifold loss and regularizers. Although PARIS learns a motion prior and LEIA implicitly performs state interpolation, both methods perform similarly for single-part objects. However, our approach excels with multi-part objects, outperforming PARIS significantly due to its flexibility in handling various motion types and articulations without constraints.}
    \label{tab:results}
\begin{adjustbox}{width=\columnwidth,center}
\renewcommand{\tabcolsep}{2pt} 
  \begin{tabular}{@{}llccccccccccccc@{}}
    \toprule
     &  & \multicolumn{6}{c}{Single-Part Articulation} & \multicolumn{6}{c@{}}{Multi-Part Articulation} \\
    \cmidrule(lr){3-8} \cmidrule(lr){9-14}
     & & {\color{gray}{45135}} & {\color{gray}{7128}} & {\color{gray}{10211}} & {\color{gray}{101917}} & {\color{gray}{103778}} & {\color{gray}{12085}} & {\color{gray}{44781}} & {\color{gray}{45427}} & {\color{gray}{45575}} & {\color{gray}{101297}} & {\color{gray}{7187}} & {\color{gray}{102377}} & \\
    Metrics & Methods & 
   Storage1 &
Microwave &
Laptop &
Oven &
Washer &Dishwasher & Storage2 & 
Storage3 & 
Storage4 & 
Sunglasses&
Oven  & Box &  Average \\
    \midrule
    PSNR$\uparrow$ 
    & PARIS & \textbf{28.66} & 25.94 & 24.97 & 28.13 & 34.46 & \textbf{27.24} & 26.35 & 24.41& 25.73 & 32.33 & 29.30 & 26.20 & 27.81 \\
    & VanillaInt & 24.20 & 22.00 & 22.63 &  24.53 & 36.03 & 24.16 & 30.96 & 29.78 & 29.10 & 32.11 & 30.35 & 27.85 & 27.81
    \\
    & \backronym & 26.69  & \textbf{26.38} & \textbf{25.04} & \textbf{28.71} & \textbf{36.14} &26.49 & \textbf{31.07} & \textbf{29.78} & \textbf{29.80} & \textbf{35.60} & \textbf{30.80} & \textbf{28.05}  &\textbf{29.55}  \\
    \midrule
    SSIM$\uparrow$ 
    & PARIS & \textbf{0.99} & \textbf{0.97} & \textbf{0.97} & 0.97 & 0.98 & \textbf{0.96} & 0.95 & \textbf{0.95} & 0.94 &  0.96 & \textbf{0.96} & 0.93 & \textbf{0.96} \\
    & VanillaInt & 0.95  & 0.91 & 0.90 & 0.93 &\textbf{ 0.99} & 0.92 & 0.95 & \textbf{0.95} &
    0.93 & \textbf{0.97} & 0.95 & 0.93 &  0.94 
    
    \\
    & \backronym &  0.97 & 0.95 & 0.95  & \textbf{0.98} & \textbf{0.99} & 0.95  & \textbf{0.96} & \textbf{0.95} & \textbf{0.95} & \textbf{0.97} & \textbf{0.96} & \textbf{0.95} & \textbf{0.96} \\
    
    \midrule
    LPIPS$\downarrow$ 
    & PARIS & \textbf{0.02} & \textbf{0.06} & 0.19 & 0.03 & 0.03 & \textbf{0.06} & 0.05 & 0.05 & \textbf{0.04} &  \textbf{0.06} & \textbf{0.05} & \textbf{0.08} & \textbf{0.06}  \\
    & VanillaInt & 0.05 & 0.10 & 0.17  & 0.03 & 0.02 & 0.08 & \textbf{0.03} & \textbf{0.03} & 
    0.05 & 0.09 &\textbf{0.05} & 0.13 &  0.07
    \\
    & \backronym & 0.03  & 0.07 & \textbf{0.10} & \textbf{0.02} & \textbf{0.02} & 0.07 & \textbf{0.03} & \textbf{0.03} & \textbf{0.04} & 0.09 & \textbf{0.05} &0.12 &  \textbf{0.06} \\
    \midrule
    CD $\downarrow$ 
    & PARIS & \textbf{0.18} & 0.08 & 0.12 &  0.75 & 0.06 & 0.45 & 0.62 & 0.42 & 0.60 & \textbf{0.20} & 0.99  & 0.91 & 0.45 \\
    
    & VanillaInt & 0.24 & 0.07 & \textbf{0.06} & \textbf{0.10} & 0.06 & \textbf{0.20} & \textbf{0.48} & 0.47 & 0.46 & 0.97 & 0.50  &0.83 & 0.37 \\
    
    & \backronym & 0.29 & \textbf{0.06}  & 0.36 & \textbf{0.10} & \textbf{0.05} & 0.27 & 0.52 & \textbf{0.38} & \textbf{0.41}  &0.96 &\textbf{0.35} & \textbf{0.62} & \textbf{0.36}\\
    \bottomrule
  \end{tabular}
  \end{adjustbox}
\end{table}

\noindent\textbf{Baselines.}
The closest prior work with a setup similar to ours is PARIS, which takes in multi-view images of objects in two states and disentangles the static and moving part of the object. While we run experiments with four input states, we don't focus on learning a defined articulation for the object, instead we are able to recognize any arbitrary motion and combinations of motions given relevant states. We emphasize on the following differences:
\begin{enumerate}[topsep=0cm]
    \item PARIS uses separate models to train objects that have rotation and translation motion, respectively. While they have a method to estimate the type of motion, once determined, it is necessary to train on the appropriate model.
    \item As PARIS also learns motion parameters for the articulation, given their pipeline of disentanglement of the static and moving parts of the object, this restricts the learning of motion parameters if there are multiple parts in the object, moving differently.
    \item \backronym is motion-prior free, and we have a universal architecture that implicitly learns intermediate states, so we are able to train with multiple types of motion occurring in the same object, with no change in the code.
\end{enumerate}
For comparing with PARIS, we run authors' original source code with our dataset, using the hyperparameters provided by the authors and training until convergence. We test our work and PARIS with the appearance quality of the reconstructed image of the interpolated state. We also set up a simple baseline that does just vanilla interpolation between the learnt latent embeddings, without enforcing any structure or constraints on them. We call this baseline VanillaInt and compare it with \backronym, which has been finetuned to do interpolation with the addition of the manifold loss and the depth and occlusion regularization for denoising the resulting output.

\begin{figure}[t!]
    \centering
    \includegraphics[trim=3.5cm 2.5cm 5cm 1.82cm, clip, width=0.9\linewidth]{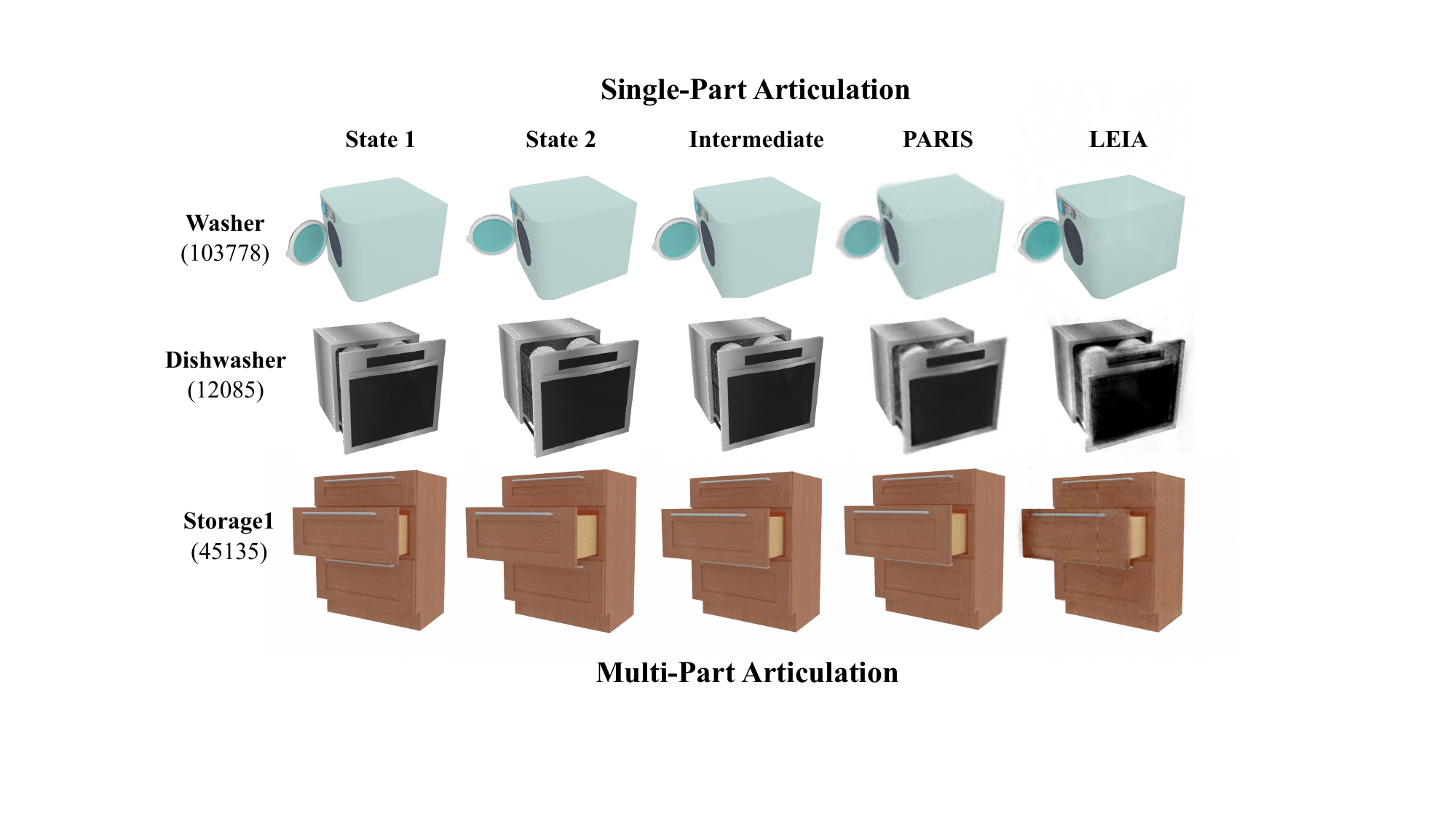}
    \centering
    \includegraphics[trim=3.5cm 0cm 5cm 0cm, clip, width=0.9\linewidth]{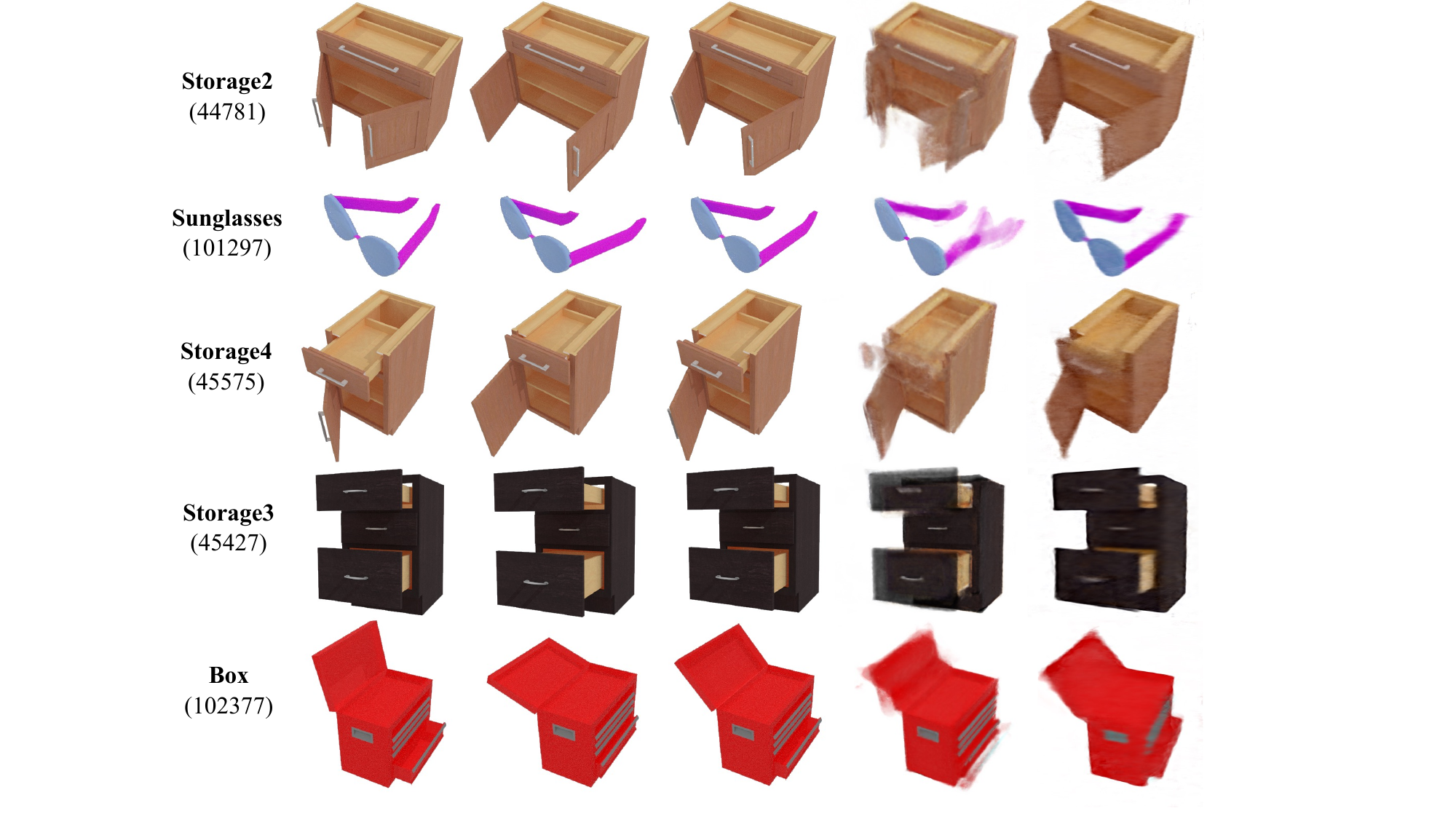}
    \caption{\textbf{Qualitative Results.} We show results of PARIS and \backronym for reconstructing the unseen intermediate state, for both single and multiple articulations. We see that PARIS especially fails when there are two parts of the object moving differently, as the motion parameters are not registered correctly. \backronym handles this case successfully as it is not dependent on part disentanglement to identify and register articulation. \backronym also performs comparable to PARIS for single-part articulation, despite not having a dedicated model for the motion or part disentanglement.}
    \label{fig:qualitative}
    
\end{figure}
\noindent\textbf{Quantitative Metrics.} We use three appearance quality metrics, PSNR (Peak Signal-to-Noise Ratio), SSIM (Structural Similarity Index), and LPIPS (Learned Perceptual Image Patch Similarity). These are computed between reconstructed images of the interpolated state, and the unseen ground truth of the state from our dataset. We also report the Chamfer Distance to measure quality of 3D reconstruction by doing point sampling.

\subsection{Main Results}
\begin{figure}[t!] 
    \centering
    \includegraphics[trim=0cm 9cm 2cm 3cm, clip, width=0.9\linewidth]{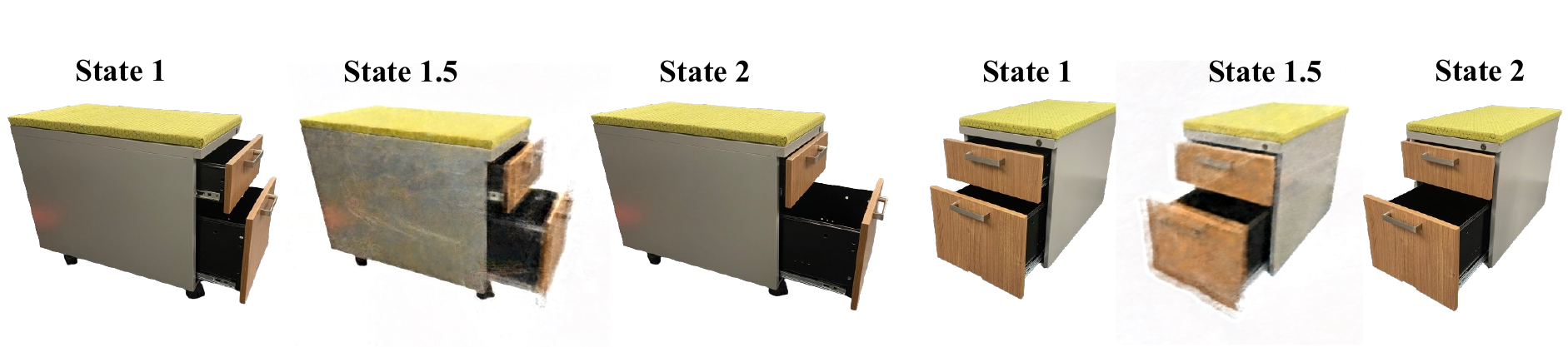}
    \caption{\textbf{Real World Results.} \backronym is able to faithfully interpolate and reconstruct between two states of images from our real world data, proving its ability to generalize and work in an in-the-wild setting.}
    \label{fig:rw_results}
\end{figure}

\begin{figure*}[t!]
    \centering
    \includegraphics[trim=0cm 1cm 1cm 0cm, clip, width=\linewidth]{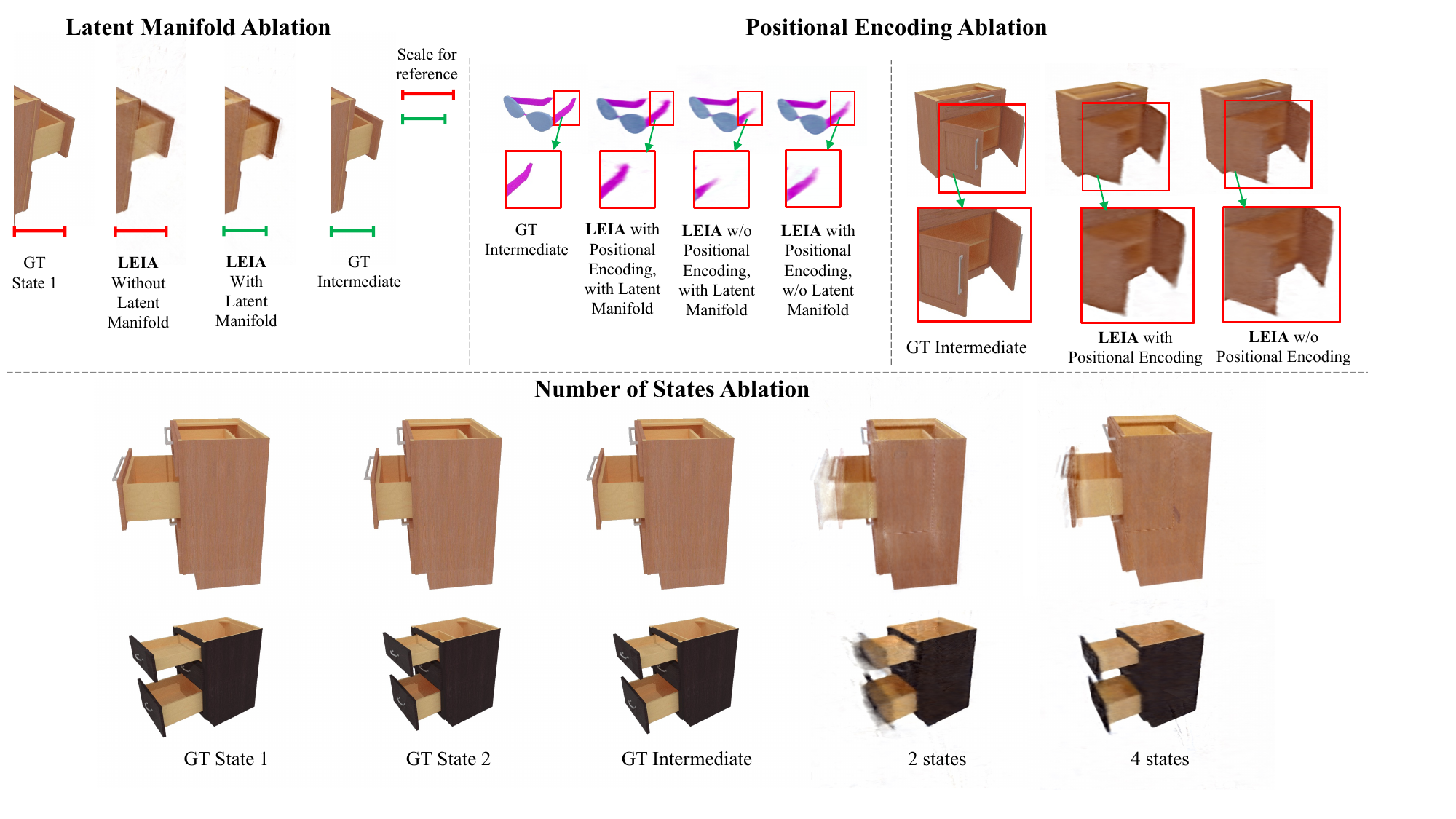}
    \caption{\textbf{Qualitative Results of Ablations.} We show some qualitative results of using latent manifold loss, positional encoding and varying the number of states. Using the manifold loss prevents us from overfitting to the extreme states, and using four states help us interpolate with a lot of clarity. Positional encoding helps add missing information for thin parts, but for large objects it doesn't help much.}
    \label{fig:ablation_qualitative}
\end{figure*}

\noindent \textbf{Novel State Synthesis.}
We show the qualitative and quantitative results of interpolating our learned latent embeddings corresponding to the start and end states, in \Cref{tab:results} and \Cref{fig:qualitative}.
Our results show that the latent embeddings can be linearly interpolated, and have a structure in the latent space. We can generate any number of states between the start and the end state by doing a weighted combination of the latents representing the start and the end states. For training, we choose four states, using the two additional states compared to PARIS to enforce structure in the latent space with the latent manifold loss, thereby making them amenable to interpolation. This helps us extend our formulation to multiple articulations without restriction. We also show results of our baseline VanillaInt and show that the numbers are boosted with the addition of the latent manifold loss and regularizers. For single-part objects, we perform comparably to PARIS despite having no part disentanglement and no dedicated model to represent the exact type of motion. For multiple-part articulations, we beat PARIS across all objects, as shown in \Cref{fig:qualitative}. Our comparable performance in 3D reconstruction shows that \backronym's learnt view-invariant latent embeddings are powerful in preserving spatio-temporal consistency. We also show that our method can work in the real world setting, in \Cref{fig:rw_results}. Ablations are shown for number of states, latent manifold loss, and depth and occlusion regularizations in \Cref{tab:ablations} and \Cref{fig:ablation_qualitative}.

\noindent \textbf{Analysis on Disentanglement of Joints.} To check separability of the latent space, we train \backronym with multi-view images across multiple states, captured from a video that showed three parts of an object \texttt{Storage1}, move one after the other. We do a t-SNE based dimensionality reduction on the embeddings and plot them, shown in Figure \ref{fig:tsne}. The latents exhibit clear separability, with the motion of each joint represented in a smooth trajectory, moving outward as the respective drawer in the object moves away from the starting ``closed'' state. This shows that the separated latent space is an indicator of multiple joints, positioning us to be capable of learning representations of multiple articulations.

\begin{table}[!t]
    \footnotesize
\begin{minipage}{0.45\textwidth}
  \includegraphics[trim=0cm 0cm 14cm 0cm, clip, width=\linewidth]{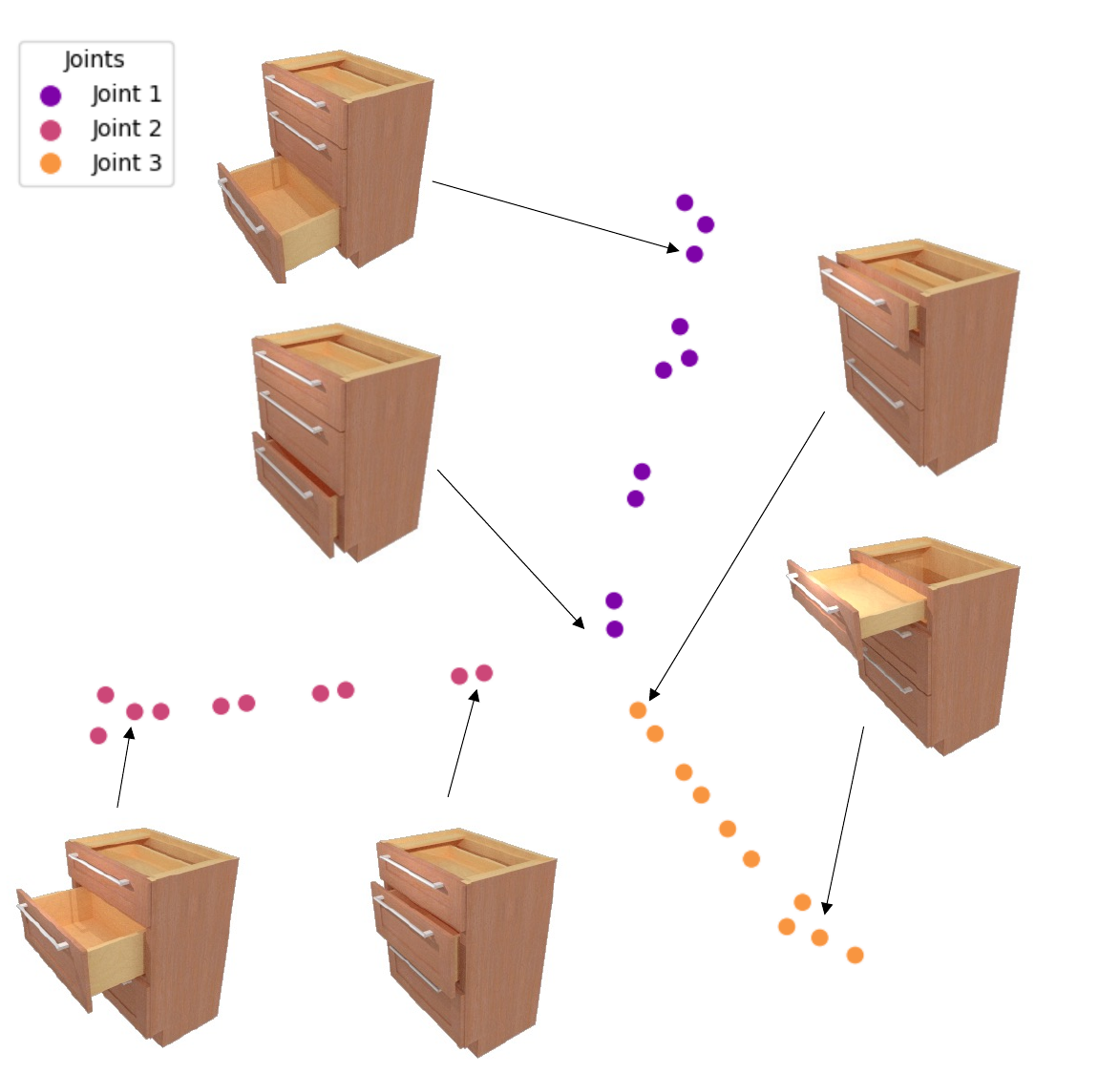}
  \captionof{figure}{\footnotesize \textbf{t-SNE plot.} After dimensionality reduction on jointly-learned state embeddings of an object with different moving parts. Our learned representations are separated and follow a smooth trajectory for each of the moving parts.}
    \label{fig:tsne}  
    \end{minipage}
    \hfill
    \begin{minipage}{0.5\textwidth}
        \centering
        \caption{\footnotesize \textbf{Ablations.} Our ablations reveal that latent manifold loss, depth, and occlusion regularization enhance \backronym's visual metrics. Opting for four states improved latent structure with the manifold loss.}
        \label{tab:ablations}
\renewcommand{\tabcolsep}{5pt}
\begin{adjustbox}{width=0.95\columnwidth,center}
\begin{tabular}{@{}lccc@{}}
\toprule
\textbf{Latent Manifold Loss} &  \textbf{PSNR} & \textbf{SSIM} & \textbf{LPIPS} \\
\midrule
with & \textbf{29.40} & \textbf{0.95} & \textbf{0.05} \\
without & 28.54 & 0.94 & 0.06\\
\midrule

\textbf{Depth Regularization} &  \textbf{PSNR} & \textbf{SSIM} & \textbf{LPIPS} \\
\midrule
with & \textbf{29.63} & \textbf{0.96} & \textbf{0.05} \\
without & 26.93 &0.93 &0.07\\
\midrule

\textbf{Occlusion Regularization} &  \textbf{PSNR} & \textbf{SSIM} & \textbf{LPIPS} \\
\midrule
with &\textbf{29.64}& \textbf{0.95} & \textbf{0.05}   \\
without & 28.64 & 0.95 & 0.06 \\

\midrule
\textbf{Positional encoding} &  \textbf{PSNR} & \textbf{SSIM} & \textbf{LPIPS} \\
\midrule
with &  27.11 & 0.94 & 0.06 \\
without & \textbf{28.48} & \textbf{0.95} & \textbf{0.06} \\
\midrule
\textbf{Number of States} &  \textbf{PSNR} & \textbf{SSIM} & \textbf{LPIPS} \\
\midrule
2 &  28.04 & 0.95 & 0.06 \\
4 & \textbf{29.69} & \textbf{0.96} & \textbf{0.05} \\
\bottomrule
\end{tabular}
\end{adjustbox}
\hfill 
\end{minipage}
\end{table}

\subsection{Ablation Analysis}
We perform ablations to tune various design choices, with results shown in \Cref{tab:ablations} and \Cref{fig:ablation_qualitative}. The first experiment we performed was testing out the impact of the latent manifold loss. We notice qualitatively that without the manifold loss, our latent space lacked structure and the linear relationship between the embeddings was not captured. As shown in \Cref{fig:ablation_qualitative}, the intermediate state overfits to the extreme state without the latent manifold while the addition of the loss enables it to accurately capture the intermediate state. This figure also shows the effect of positional encoding which adds detail when the latent embedding fails to capture it, as evidenced in the results of \texttt{sunglasses}. This can happen for thin parts of the object that may not be not well-captured across camera views. Notably, the inverse happens in the case of large objects, where the positional encoding adds extra noise to the reconstruction, resulting in over-smoothing. We also show how adding just two more intermediate states in \backronym makes a huge difference in reconstruction results for both single and multi-part objects, enabling us to scale and be flexible. Quantitative numbers for depth and occlusion regularization ablation are shown in \Cref{tab:ablations}. However, they didn't universally help in the cases where the latent embedding already learned the state representation well enough but reduced noise when it appeared.

\section{Limitations}
\begin{figure} [t!]
    \centering
    \includegraphics[trim=0cm 13cm 0cm 0cm, clip, width=\linewidth]{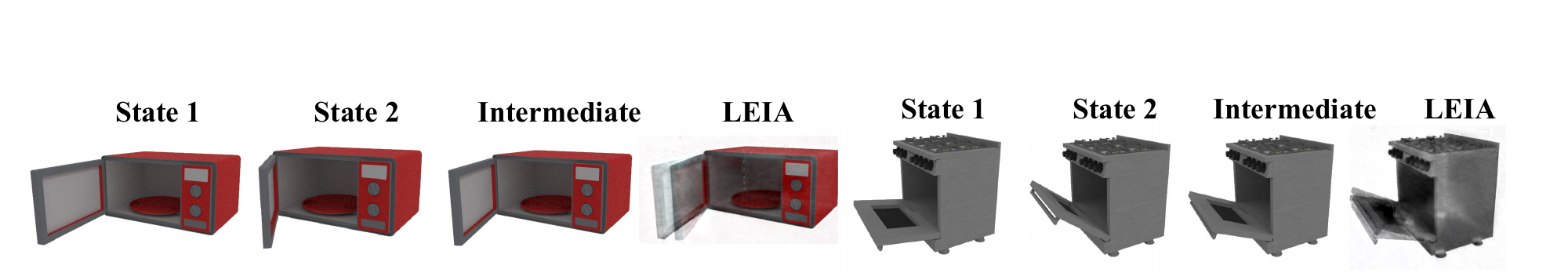}
    \caption{\textbf{Failure cases.} Our model fails at at reconstructing the geometry correctly at camera positions where the two states we interpolate between have a change in the visible shape, like these examples where the microwave and oven being closed shows a deformation in the figure as compared to the open state, as the relationship between the motion is not easily captured in the latent due to the structure change.}
    \label{fig:failure}
\end{figure}

While \backronym works in achieving good quality reconstruction of intermediate states, our latent embeddings do not yet ensure accurate physical consistency in the motion of the intermediates, which is a tradeoff we chose while opting for not decoupling the object and learning the motion parameters for the moveable part. This allowed us to scale our approach to interpolate between multiple moving parts of the object, which would be more representative of the real world where common objects can be articulated in multiple ways.
\backronym also struggles when there is severe occlusion, as referenced by Figure \ref{fig:failure}, as the occlusion occurs at gaps in the learnt embeddings between two states.

\section{Conclusion}
In this work, we present \backronym, a method capable of successfully interpolating between two discrete articulation states of a deformable object with moving parts. \backronym handles multiple joints, including cases where moving parts are separated by static regions. It outperforms existing methods, especially when multiple parts move independently. The learned latent embeddings are view-invariant and separable, demonstrating \backronym's scalability and flexibility. We conducted comprehensive evaluations on synthetic and real data to investigate inherent challenges. Despite advancements, significant occlusion scenarios remain challenging. We hope future research builds upon our work.

\section*{Acknowledgements}
This work was partially supported by NSF CAREER Award (\#2238769) to Abhinav Shrivastava, and IARPA via Department of Interior/Interior Business Center (DOI/IBC) contract number 140D0423C0076. The U.S. Government is authorized to reproduce and distribute reprints for Governmental purposes notwithstanding any copyright annotation thereon. The views and conclusions contained herein are those of the authors and should not be interpreted as necessarily representing the official policies or endorsements, either expressed or implied, of NSF, IARPA, DOI/IBC, or the U.S. Government.

%
%

\end{document}